\documentclass[]{bytedance_seed}

\usepackage[toc,page,header]{appendix}
\usepackage{comment}

\usepackage{minitoc}
\usepackage{algorithm}      
\usepackage{algpseudocode}  
\usepackage{graphicx}
\usepackage{subcaption}
\usepackage{xspace}
\usepackage{multirow}
\usepackage[table]{xcolor}
\usepackage[dvipsnames]{xcolor}

\newcommand{\para}[1]{\noindent \textbf{#1}}
\newcommand{\done}[1]{\textcolor{red}{{[DONE]}}}
\usepackage{amssymb}
\usepackage{booktabs}
\usepackage{adjustbox}
\usepackage{accsupp}
\usepackage{transparent}
\definecolor{ShortText}{RGB}{144, 216, 212}  
\definecolor{LongText}{RGB}{128, 200, 195}   
\definecolor{Visual}{RGB}{210, 180, 220}     
\definecolor{bellow}{RGB}{229, 181, 89}  

\newcommand{\bagel}{%
  \raisebox{-0.2ex}{\includegraphics[height=1em]{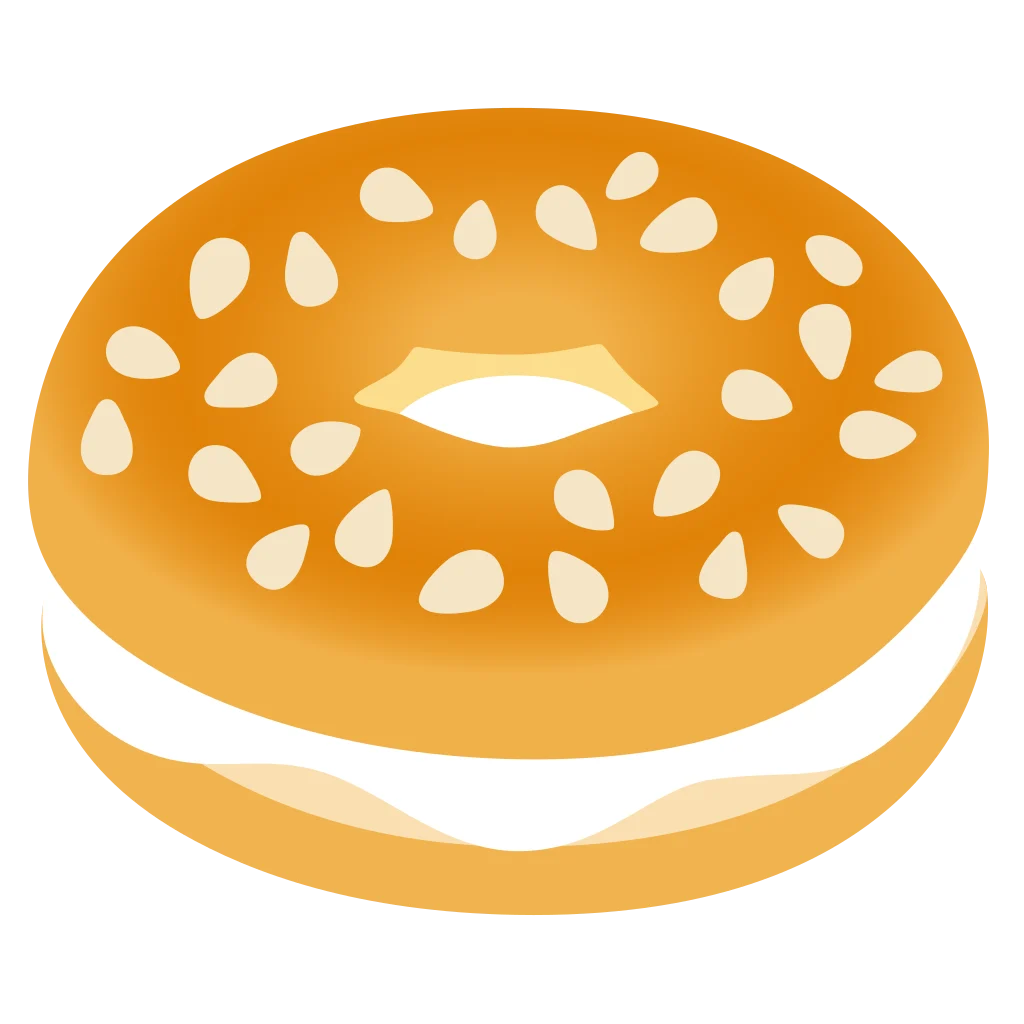}}%
}

\newcommand{\method}{%
  \mbox{%
    \rlap{\textcolor{white}{O}}%
    \bagel\textcolor{bellow}{mni}%
  }%
  \xspace%
}

\title{Context Unrolling in \method Models}

\author[*,\dagger]{Ceyuan Yang}
\author[*]{Zhijie Lin}
\author[*]{Yang Zhao}
\author[*]{Fei Xiao}
\author[*]{Hao He}
\author[*]{Qi Zhao} 
\author[]{\\ Chaorui Deng}
\author[]{Kunchang Li}
\author[]{Zihan Ding}
\author[]{Yuwei Guo}
\author[]{Fuyun Wang}
\author[]{\\ Fangqi Zhu}
\author[]{Xiaonan Nie}
\author[]{Shenhan Zhu}
\author[]{Shanchuan Lin}
\author[]{\\ Hongsheng Li}
\author[]{Weiling Huang}
\author[\dagger]{Guang Shi}
\author[]{Haoqi Fan}

\contribution[*]{Equal contribution}
\contribution[\dagger]{Corresponding authors}

\abstract{
We present \method,  a unified multimodal model natively trained on diverse modalities, including text, images, videos, 3D geometry, and hidden representations. We find that such training enables \textit{Context Unrolling}, where the model explicitly reasons across multiple modal representations before producing predictions. This process enables the model to aggregate complementary information across heterogeneous modalities, facilitating a more faithful approximation of the shared multimodal knowledge manifold and improving downstream reasoning fidelity. As a result, \method achieves strong performance on both multimodal generation and understanding benchmarks, while demonstrating advanced multimodal reasoning capabilities, including in-context generation of text, image, video, and 3D geometry.
}

\date{\today}
\checkdata[Project Page]{\url{https://omni-model.com/}}

\begin{document}
\maketitle


\section{Introduction}

In the native multimodal regime, unified multimodal models \cite{openai2024gpt4o,google2024gemini15,shukor2023unival,gong2025mingomni,mizrahi2023fourm,zhan2024anygpt,deng2025bagel} learn a world knowledge manifold by taking inputs and predicting outputs across multiple modalities. 
Each modality provides only a partial and biased view of this multimodal manifold, capturing complementary aspects of world knowledge. 
We argue that native multimodal models can explicitly benefit from \textit{Context Unrolling} - reasoning across these heterogeneous modal projections to recover a more complete approximation of the shared multimodal manifold. 
We observe that prediction quality improves as models perform Context Unrolling across more modalities, integrating information from a broader set of modalities before producing outputs.
By structuring reasoning across different modal projections, the model forms a more faithful representation of the multimodal manifold, producing outputs that are more coherent, faithful, and semantically accurate.

\begin{figure}[t]
    \centering
    \includegraphics[width=1.0\linewidth]{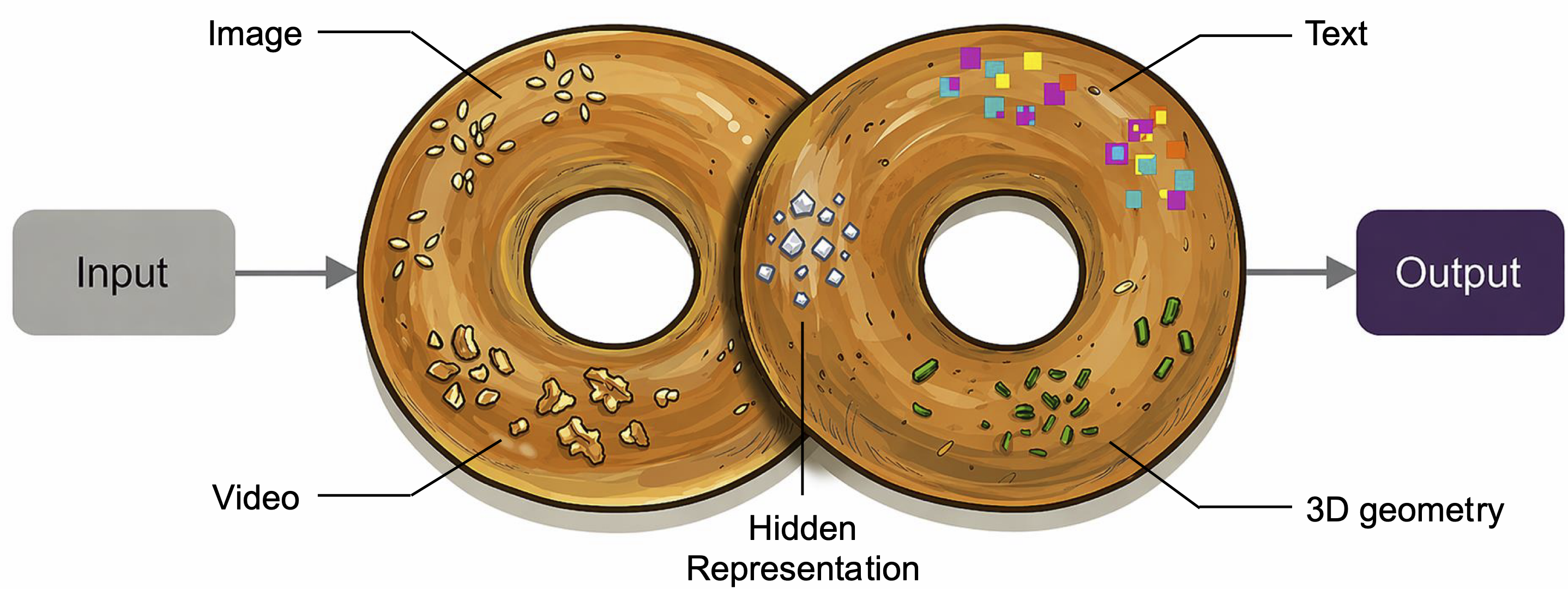}
    \caption{Given an arbitrary task, \method selectively activates task-relevant contexts from a heterogeneous context pool—spanning text, image, video, 3D geometry, and beyond—into a shared workspace before producing predictions. This mechanism enables the model to aggregate complementary information across modalities, improving downstream reasoning and generation fidelity.}
    \label{fig:context_unrolling}
\end{figure}

To realize this vision, we follow the design philosophy of BAGEL~\cite{deng2025bagel}, and expand training modalities from image–text pairs to a broader set of modalities including text, images, videos, 3D geometry, and hidden visual representations. The expanded set of modalities provides complementary projections of world knowledge, capturing information such as pixel appearance, spatial-temporal structure, camera transformations, depth, physical dynamics, semantic abstraction, and dense conversational reasoning, which are essential for learning grounded world knowledge at scale. Building upon the interleaved data paradigm introduced in BAGEL~\cite{deng2025bagel}, we further incorporate reasoning-oriented multimodal content to encourage structured cross-modal reasoning during training. This enables generation to leverage not only short textual reasoning but also long-form structured descriptions with dense attributes, spatial grounding, and geometric constraints such as depth maps and camera transformations. We additionally introduce a hidden reasoning space as a dedicated latent representational space to support latent multimodal reasoning. Through this design, the unified model learns to reason across heterogeneous modal projections while producing coherent multimodal predictions, enabling richer cross-modal interaction and strengthening native multimodal learning capabilities.

We finally introduce \method, a multimodal foundation model that supports any-to-any multimodal learning. Built upon BAGEL~\cite{deng2025bagel}, \method perform context unrolling across modalities to enable cross-modal reasoning. It contains 3B active parameters and adopts a mixture-of-experts architecture. 
Compared with existing models that typically map multimodal inputs to a single output modality, \method unifies image, video, and text for understanding, generation, and editing within a single architecture, achieving competitive or state-of-the-art performance across a wide range of benchmarks.

The model achieves competitive or superior performance compared to leading open-source vision-language models, including Qwen3-VL~\cite{Qwen3-VL} and InternVL3.5~\cite{wang2025internvl3}, across standard multimodal understanding benchmarks. It also outperforms strong public image generators such as Z-Image~\cite{cai2025z}, Flux~\cite{lipman2022flow}, and Qwen-Image~\cite{wu2025qwen} on the GenEval2 benchmark, and demonstrates consistently stronger qualitative performance on classical image editing tasks compared to Step1X~\cite{liu2025step1x}, Qwen-Image-Edit~\cite{wu2025qwen}, and Z-Image-Edit~\cite{cai2025z}.
Beyond static image manipulation, \method extends unified multimodal modeling to video generation and editing, demonstrating strong semantic instruction-following capabilities in temporal visual reasoning and achieving superior qualitative performance compared to Wan2.1~\cite{wan2025wan} and Hunyuan~\cite{kong2412hunyuanvideo} on video generation and editing tasks. 
In 3D geometry, \method also achieves comparable performances to VGGT~\cite{wang2025vggt} in camera estimation and to Depth-Anything 3~\cite{depthanything3} in depth estimation. By incorporating video as a native modality, \method provides a unified framework for text, image, video and 3D geometry within a single model architecture.

With modality-scaled pretraining in \method, we observe an emerging capability that we term \textit{Context Unrolling}. Given an arbitrary task, a native multimodal model develops the ability to unroll its internal “thinking” across heterogeneous modalities. Each modality can be viewed as a projection of a shared latent world knowledge space, where modality-specific representations provide complementary evidence for completing the reasoning process. Through this mechanism, the model dynamically integrates and selects modality-specific information, enabling more comprehensive and structured inference. The emergent \textit{Context Unrolling} capabilities are supported not only by test-time improvements across diverse public benchmarks, including visual understanding and visual generation tasks, but also by probing tasks such as depth estimation and spatial reasoning. These findings suggest that native unified multimodal pretraining naturally encourages cross-modal reasoning and externalization of world knowledge through modality-specific contexts. This ability substantially improves inference-time reasoning quality and demonstrates the potential of unified multimodal models as foundational reasoning systems.

\section{Context Unrolling}

\begin{figure}[t]
    \centering
    \includegraphics[width=0.32\linewidth]{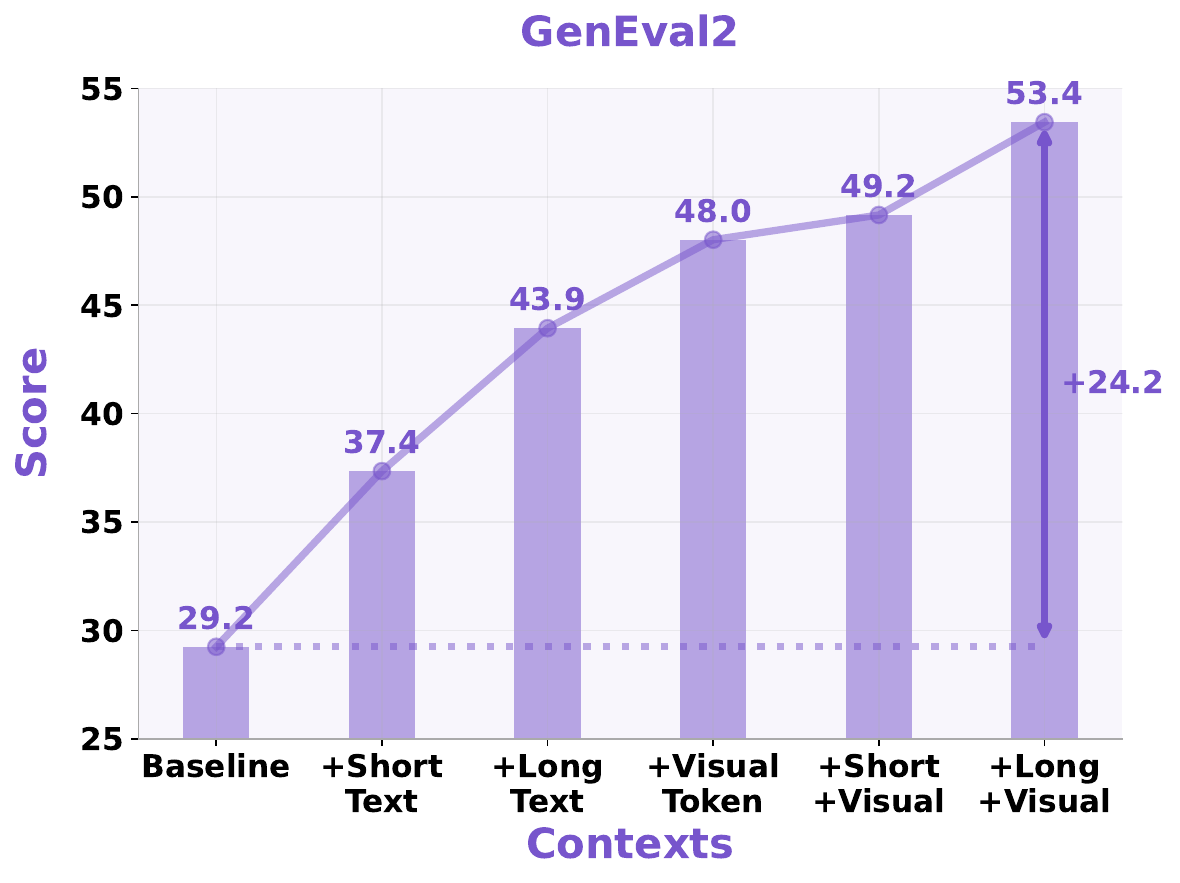}
    \includegraphics[width=0.32\linewidth]{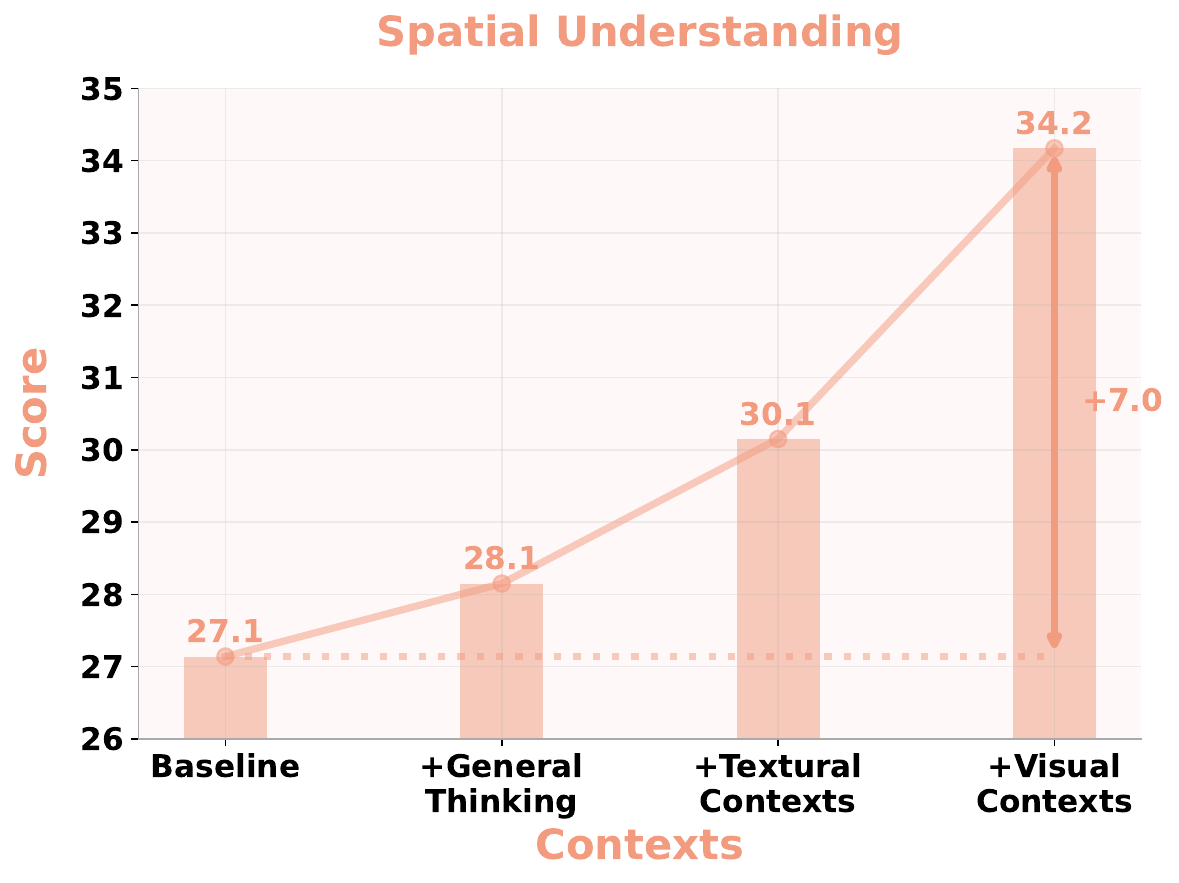}
    \includegraphics[width=0.32
    \linewidth]{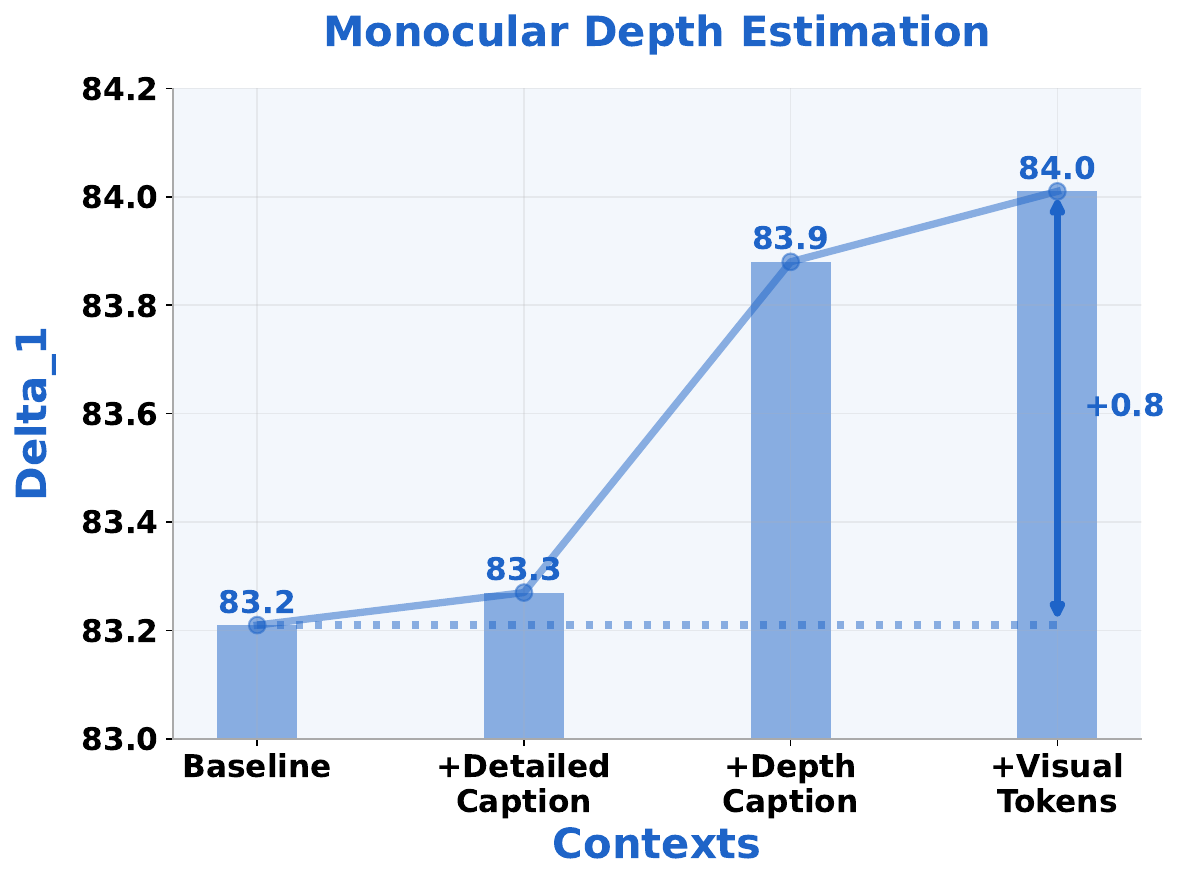}
    \caption{\textbf{Context Unrolling.} Multi-model model benefits from multi-granularity contexts. With more fine-grained textual specifications and visual tokens that carry strong structural signals, generation obtains significant gains. Spatial understanding can be improved via 3D geometry and visual imagination. Monocular depth estimation can be promoted by both textural and visual reasoning.}
    \label{fig:context_unrolling}
\end{figure}

A unified model should not be understood as a ``multi-task container'' that merely places multimodal understanding, image/video generation, and 3D geometry behind a shared backbone. 
The more substantive benefit is unified context unrolling: unification turns each capability into an atomic primitive that can be invoked, composed, and written back into a shared workspace (context), which then conditions subsequent computation. 
Under this view, tasks are not isolated endpoints; they are operators over a growing context—a context that can include (i) fine-grained textual reasoning (chain-of-thoughts), (ii) structured intermediate representations such as 
visual tokens rolled out by the model itself, and (iii) geometric cues (camera pose, depth, and view synthesis). 
As the model expands this workspace, downstream predictions become better constrained (less ambiguous), more structure-preserving, and more cross-task consistent.

Concretely, we model inference as iterative context construction followed by context-conditioned decoding:
\begin{equation}
C_{t+1}=C_t \oplus \phi_t(x, C_t), \qquad
y=\psi\!\left(x \mid C_T\right),
\end{equation}
where $x$ denotes the multimodal inputs, $\phi_t$ are atomic primitives (e.g., ``describe,'' ``predict pose,'' ``roll out visual tokens,'' ``synthesize a novel view,'' ``estimate depth''), and $\oplus$ denotes context composition. This reframes ``unification'' as a mechanism that scales context—not only in length, but in structure and utility for downstream decisions. \Cref{fig:context_unrolling} presents the overall results with context unrolling, which are further unfolded in the following content.

\subsection{Visual Understanding}
Here, context unrolling primarily occurs through the spontaneous ``think'' / CoT-style textual rollout, which enriches the latent workspace with finer semantic decompositions before producing the final answer. The resulting context improves compositional reasoning even when the task itself is ``standard'' VLM. We validate the performance of \method on a downsampled visual understanding benchmark. As expected in \Cref{tab:vlm_thinking}, visual understanding can be improved with thinking context on various dimensions. 
\begin{table}[t]
\centering
\caption{\textbf{Understanding with Self-Thinking.} We validate the performance on an internal donwsampled benchmark.}
\label{tab:vlm_thinking}
\renewcommand{\arraystretch}{1.0}
\resizebox{\textwidth}{!}{%
\begin{tabular}{l|ccccccccccc}
\toprule
 Context & \textbf{BLINK}$_{\uparrow}$ & \textbf{MMStar}$_{\uparrow}$ & \textbf{MMBench-V11}$_{\uparrow}$ & \textbf{SimpleVQA}$_{\uparrow}$ & \textbf{AI2D}$_{\uparrow}$ & \textbf{Chartqa}$_{\uparrow}$ & \textbf{Docvqa}$_{\uparrow}$ & \textbf{HallusionBench}$_{\uparrow}$ & \textbf{Erqa}$_{\uparrow}$ & \textbf{MMSI}$_{\uparrow}$\\
\midrule
\method & 60.8 & 59.4 & 76.2 & 50.4 & 90.2 & 85.5 &93.5 & 69.6 & 41.5 & 31.5\\

+~\textcolor{LongText}{thinking} & 61.6 & 66.5 & 77.1 & 51.4 & 92.3 & 88.0 &94.0 & 71.3 & 44.5 & 32.6\\
\bottomrule
\end{tabular}
}
\end{table}

\subsection{Visual Generation}


\begin{table}[t]
\centering
\caption{\textbf{Text-to-Image Generation on Benchmarks.} GenEval-2 (left) and inhouse evaluation (right) provide a comprehensive validation where TIFA$_{GM}$ measures prompt-level correctness. Atomicity is also reported to evaluate robustness under increasing prompt compositionality.  Regarding contexts, \textcolor{Visual}{visual} contexts denote the discrete visual tokens. \textcolor{ShortText}{Short} and \textcolor{LongText}{long} texts are also derived from our model itself, providing fine-grained detailed descriptions. The \textcolor{LongText}{oracle} text is produced by Gemini-3 Pro in the zero-shot manner, plus \textcolor{Visual}{oracle visual} contexts, to see how far our generation can go.}
\label{tab:text_thinking_geneval2}
\resizebox{\textwidth}{!}{
\begin{tabular}{l|c|ccccc}
\toprule
\textbf{Context} & \textbf{TIFA$_{GM}$} & \textbf{Object} & \textbf{Attribute} & \textbf{Count} & \textbf{Position} & \textbf{Verb} \\
\midrule \method & 29.25 & 91.64 & 90.00 & 52.03 & 77.67 & 26.25 \\
\midrule
+~\textcolor{ShortText}{short} & 37.35 & 93.18 & 92.45 & 60.14& 76.92 & 38.83 \\
+~\textcolor{LongText}{long} & 43.94 & 91.86 & 91.13 & 67.03 & 77.03 & 38.31 \\
+~\textcolor{Visual}{visual} & 48.02 & 94.42 & 92.96 & 66.92 & 79.28 & 53.96 \\
\midrule
+~\textcolor{ShortText}{short}~and~\textcolor{Visual}{visual}      & 49.16 & 93.13 & 92.68 & 68.36 & 76.83 & 43.34 \\
+~\textcolor{LongText}{long}~and~\textcolor{Visual}{visual}        & 53.44 & 92.34 & 92.32 & 72.98 & 80.23 & 42.81 \\
\midrule
+~\textcolor{LongText}{oracle}                  & 52.20 & 95.72 & 87.35 & 67.91 & 91.69 & 43.31 \\
+~\textcolor{LongText}{oracle}~and~\textcolor{Visual}{visual}     & 57.21 & 94.77 & 97.89 & 69.47 & 90.64 & 56.00 \\
\bottomrule
\end{tabular}

\begin{tabular}{|c|ccc}
\toprule
 \textbf{Overall} & \textbf{Object} & \textbf{Attribute} & \textbf{Relation} \\
\midrule 
 0.56 & 0.60 & 0.58 & 0.53 \\
\midrule
 0.59 & 0.68 & 0.61 & 0.58 \\
 0.61 & 0.72 & 0.64 & 0.57 \\
 0.61 & 0.70 & 0.65 & 0.62 \\
\midrule
 0.64 & 0.74 & 0.65 & 0.62 \\
 0.66 & 0.74 & 0.69 & 0.64 \\
\midrule
 0.71 & 0.75 & 0.73 & 0.74 \\
0.73 & 0.79 & 0.76 & 0.73 \\
\bottomrule
\end{tabular}
}
\end{table}

Visual generation particularly benefits from multi-granularity context to reduce the inherent ambiguity of mapping language to images. From the unified context unrolling perspective, image synthesis is not an isolated endpoint; instead, it is a context-conditioned decoding process where upstream atomic primitives (e.g., textual reasoning and structural token rollouts) can be invoked to construct a richer workspace before pixel synthesis.

To isolate the effect of different contexts, we focus on text-to-image (T2I) generation as an analysis pretext task, leveraging its comprehensive and standardized benchmarks.
Concretely, before synthesizing an image, \method can optionally (i) roll out more fine-grained textual specifications (attributes, counts, relations, spatial constraints) via \texttt{text-think} 
, and/or (ii) roll out 
visual tokens that carry strong structural information. Conditioning the generator on these intermediate contexts improves prompt following, object counting, and spatial/relational fidelity—illustrating that the unified model’s gain stems from unrolling usable contexts.

\para{Self-Unrolling Contexts.} 
\Cref{tab:text_thinking_geneval2} reports results on GenEval-2 and inhouse evaluation, where Soft TIFA$_{GM}$ measures prompt-level correctness. We also report atomicity, which evaluates robustness under increasing prompt compositionality. As highlighted by GenEval-2, T2I performance typically drops sharply as compositionality increases. Therefore, atomicity serves as a stress test for text following and compositional generalization. 

Following prior practices of unified models (e.g., BAGEL), we can ask the model to think before generation, producing fine-grained textual descriptions under different token budgets. In our setup, short thinking consumes around 100 tokens on average, while long thinking uses around 250 tokens and typically yields richer information, including coarse structural layouts.

Overall, richer textual context consistently improves both the aggregate performance and the performance under higher atomicity (i.e., more compositional prompts), confirming the benefit of unrolling textual context.
Moreover, using 
visual tokens only enhances generation across metrics, with particularly strong gains on counting- and action/verb-related prompts, consistent with the hypothesis that 
visual tokens inject explicit structural cues that are otherwise difficult to preserve through pure text conditioning.
Importantly, as the number of atomic requirements in the prompt increases (i.e., more complex descriptions), \method retains a moderate level of prompt-following ability, indicating improved robustness under compositional stress.
Finally, textual and visual contexts are complementary: combining \texttt{text-think} with 
visual token rollout yields the most consistent improvements, supporting our central claim that unified models excel by composing multiple atomic capabilities into a stronger shared context.

\para{Oracle Contexts.} 
While self-rollout contexts already bring significant improvements, the limited capacity of current models can introduce noise and hallucinated details, which may cap the benefit of context unrolling. We therefore perform oracle studies to estimate how far generation can be pushed under near-ground-truth contexts.

We view the rolled textual context as a form of understanding-driven prompt rewriting: it reduces uncertainty by making high-level concepts, spatial configurations, and fine-grained appearance constraints more explicit. To approximate an upper bound on textual contexts beyond the current computation scale, we use Gemini-3 Pro to provide higher-quality rewrites in a zero-shot manner, serving as an oracle textual context in this study.

\Cref{tab:text_thinking_geneval2} shows that oracle contexts yield a substantial leap in generation quality, while simultaneously exposing the gap to current self-rollout contexts—direct evidence that unrolling context (quality and structure) is a primary driver of performance.
Notably, even when Gemini-3 Pro provides strong textual contexts, adding self-rollout visual tokens further improves results, suggesting that (i) textual and visual contexts provide non-redundant constraints, and (ii) structural visual tokens remain valuable as an additional context channel. Together, these findings reinforce our thesis: the unified model improves generation by invoking and composing atomic capabilities to construct richer, more actionable context prior to decoding.

\subsection{Spatial Understanding}
\begin{figure}[!t]
    \centering
    \includegraphics[width=\linewidth]{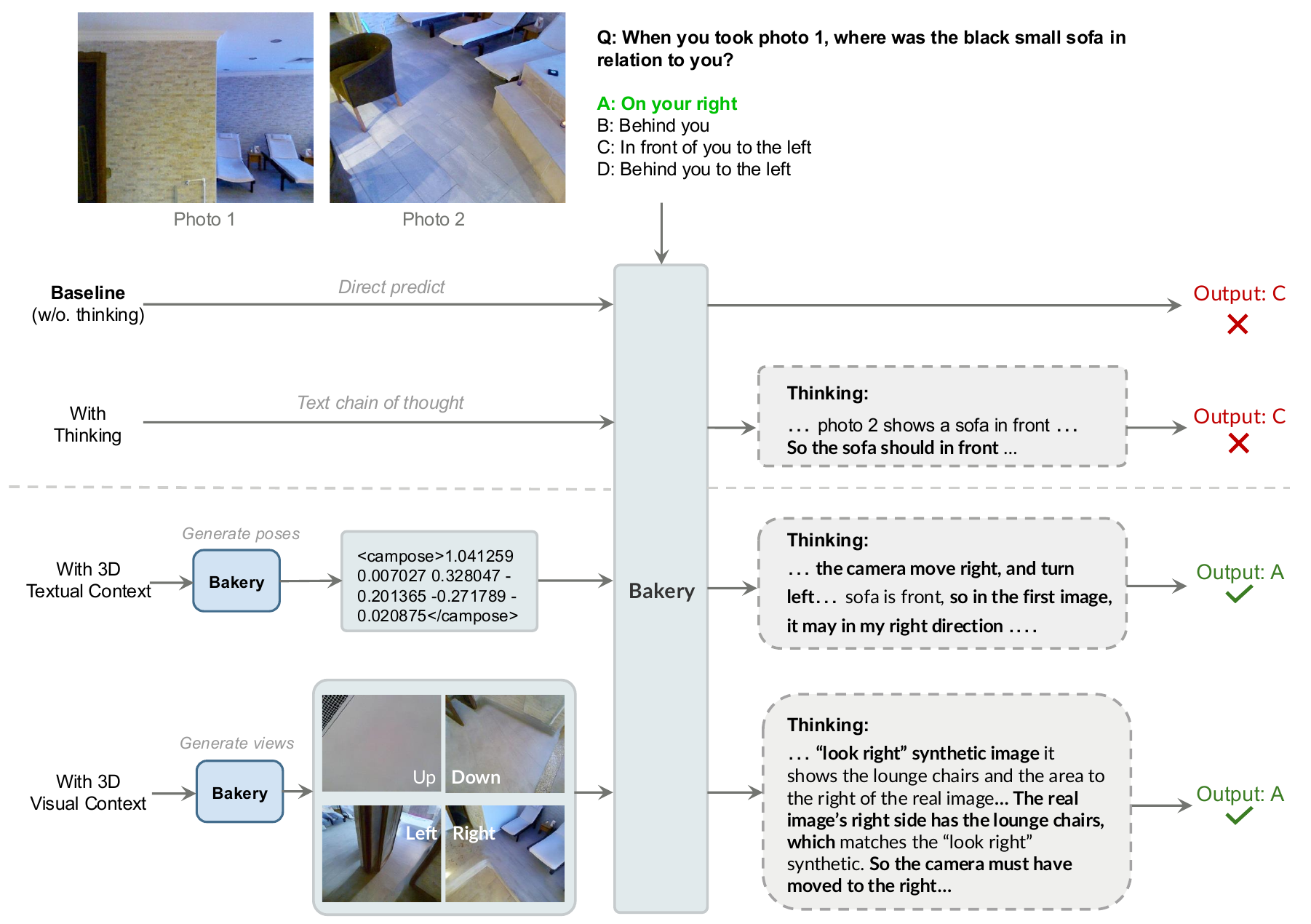}
    \caption{\textbf{Illustration of Context Unrolling on Spatial Understanding}. Given a question about an object’s relative position across two views, the baseline (direct prediction) and text-only chain-of-thought both fail. Augmenting the VLM with explicit 3D textual context (relative camera pose) or 3D visual context (synthesized views under canonical motions: up/down/left/right) enables correct prediction.}
    \label{fig:mmsi_ablation}
\end{figure}

\begin{table}[t]
\centering
\caption{\textbf{Spatial Understanding Evaluation}. Textural contexts denote the geometry-grounded text (\textit{e.g.,} camera pose estimation results). Visual contexts refer to the novel-view synthesis results as contexts if applicable. Performances are reported on a downsampled MMSI-Bench.}
\label{tab:mmsi}
\resizebox{0.7\textwidth}{!}{
\resizebox{\textwidth}{!}{
\begin{tabular}{lcccc}
\toprule
\textbf{Context} & \textbf{Overall Score} & \textbf{MSR} & \textbf{Motion} & \textbf{Positional Relationship} \\
\midrule
\method & 27.14 & 17.65 & 0.0 & 19.63 \\
\midrule
+~\textcolor{ShortText}{thinking}         & 28.15 & 17.65 & 33.33 & 30.25 \\
+~\textcolor{LongText}{textural contexts}  & 30.15 & 11.76 & 33.33 & 33.95 \\
+~\textcolor{Visual}{visual contexts}   & 34.17 & 26.47 & 33.33 & 35.80 \\
\bottomrule
\end{tabular}
}
}
\end{table}
As multimodal understanding models mature, a key question emerges: can they understand the real world in 3D? This capability, often referred to as spatial intelligence, is widely regarded as essential for physical-world operation and embodied deployment.

A key challenge is the mismatch between visual and textual modalities: the real world is visually redundant and geometrically complex, while textual reasoning is compact and abstract. Reasoning purely in freeform text often struggles with geometric ambiguities such as viewpoint changes, foreshortening, and occlusions, limiting performance on questions requiring consistent 3D interpretation across multiple views. Our unified context unrolling framework addresses this gap by incorporating 3D-related capabilities—camera pose estimation, novel view synthesis, and depth estimation—as atomic primitives that can be invoked within the reasoning loop. To validate the effectiveness of context unrolling for spatial understanding, we select 200 questions from the MMSI benchmark~\cite{yang2025mmsi} that are closely related to 3D spatial reasoning as the testbed.

\para{3D Textural Contexts: Geometry-as-Context.} 
Analogous to general visual understanding, \method can enable a \texttt{text-think} mode before answering. For spatial questions, however, the most useful intermediate signal is often explicitly geometric. We therefore allow the model to first perform camera pose estimation from the input views before thinking and answering. The estimated camera poses serves as a 3D textual context that disambiguates spatial relations across images.

As shown in \Cref{tab:mmsi}, injecting this geometry-grounded context improves MMSI accuracy and consistently outperforms baselines that rely on latent, free-form text-only reasoning, indicating that explicit 3D contexts provide actionable constraints for downstream decision making.

\para{3D Visual Contexts: Imagination-as-Context.}
Beyond textual contexts, a unified model can leverage its generative capabilities to "think with images." Before answering spatial questions, the model first synthesizes novel views around each given observation—imagining the surrounding environment from up, down, left, and right viewpoints, etc. This process enriches the visual context by completing a more comprehensive scene representation, allowing the model to reason from a near-omniscient perspective rather than being limited to the provided viewpoints. \Cref{tab:mmsi} shows that adding such NVS-derived visual contexts (+ visual contexts) yields the strongest MMSI performance. This demonstrates that generative imagination of unseen views provides more discriminative evidence for spatial reasoning than relying solely on the given observations.

In both settings, the critical point is that geometry estimation and generation are not auxiliary tasks evaluated in isolation. They function as context-producing primitives, textual (pose summaries) or visual (synthesized evidence), that are composed to scale the effective context for spatial reasoning. \Cref{fig:mmsi_ablation} provides a concrete qualitative example of this mechanism: while standard text-only reasoning fails due to geometric ambiguity, our approach successfully invokes 3D primitives to construct actionable context—either by explicitly grounding reasoning in pose data or by "imagining" intermediate visual evidence—thereby correcting the reasoning path and guiding the model to the correct answer. Namely, the gains of unified models come from unified context unrolling, i.e., the ability to invoke and compose heterogeneous capabilities to construct richer, more actionable context before making a decision.

\begin{figure}[t]
    \centering
    \includegraphics[width=\linewidth]{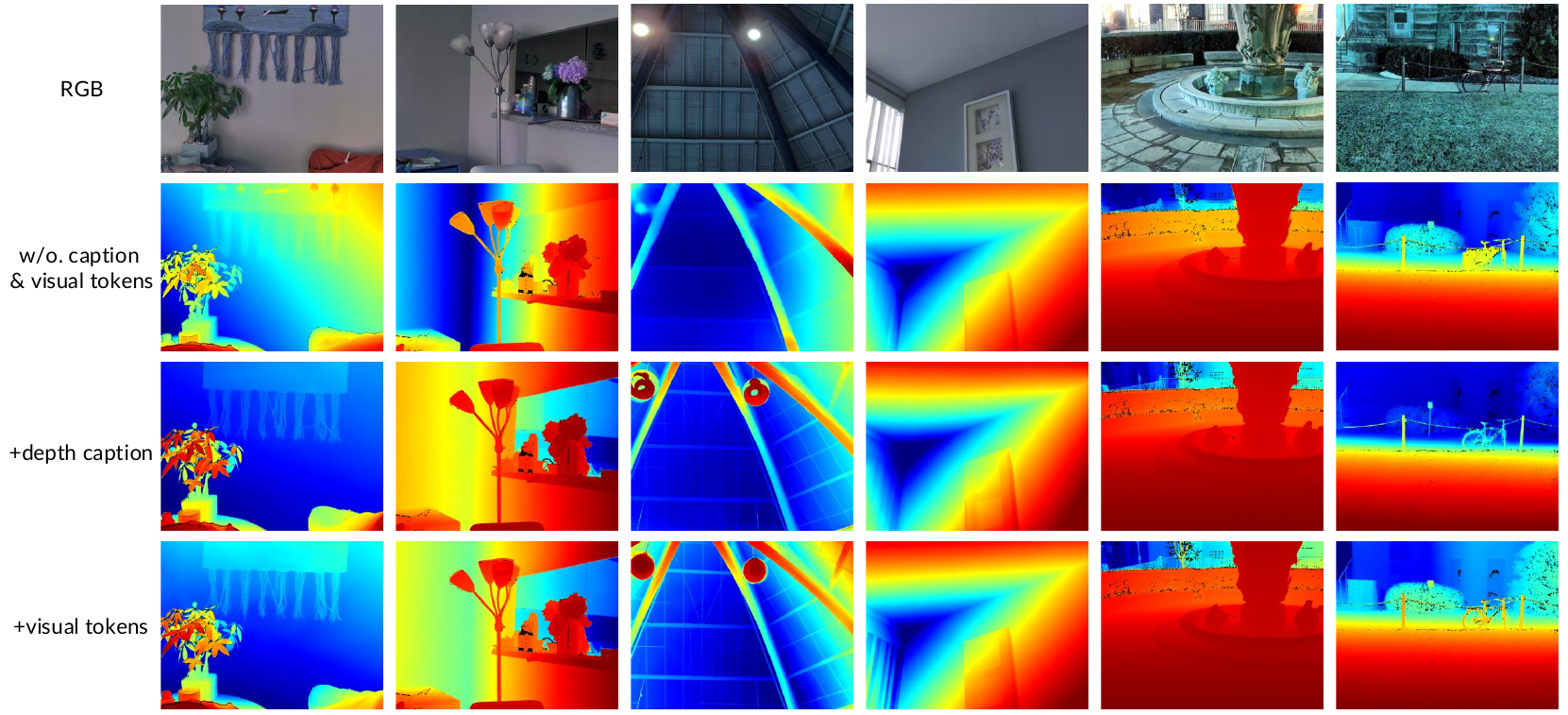}
    \caption{\textbf{Depth Estimation with Different Contexts.}}
    \label{fig:depth_ablation}
\end{figure}

\subsection{3D Geometry}

\begin{table}[t]
\centering
\caption{\textbf{Depth Estimation Errors.} Different from traditional depth estimators, \method can think before predicting. Here, the \textcolor{LongText}{detailed caption} means to produce general-purpose descriptions for the input image. The \textcolor{ShortText}{depth caption} focuses on the spatial cues. \textcolor{Visual}{Visual} contexts denote the proposed visual tokens that enhance the structural information.}
\label{tab:3d_rec}
\renewcommand{\arraystretch}{1.0}

\begin{tabular}{l|cc}
\toprule
 Context & $\delta_1 \uparrow$ & AbsRel $\downarrow$ \\
\midrule
\method & 83.21\% & 0.2028 \\
+ \textcolor{LongText}{detailed caption} & 83.27\% & 0.2029 \\
+ \textcolor{ShortText}{depth caption} & 83.88\% & 0.1988 \\
+ \textcolor{Visual}{visual contexts} & 84.01\% & 0.1970 \\
\bottomrule
\end{tabular}

\end{table}
Recent feed-forward 3D models (e.g., VGGT~\cite{wang2025vggt}) highlight the value of multi-task training, where a strong visual encoder which is often learned via self-supervision provides rich priors for geometry prediction. In this section, we use the long-standing problem of monocular depth estimation to illustrate that the main gain of unification is not simply multi-task parameter sharing, but unified context unrolling, where other capabilities become atomic primitives that supply additional constraints to 3D Geometry.

We formulate depth estimation as depth-map generation (predicting a depth image conditioned on the input RGB). This protocol is standard and has been explored by diffusion-based approaches (e.g., Marigold~\cite{ke2023repurposing}). However, purely generative formulations may lag behind state-of-the-art feed-forward regressors when used in isolation. Our approach differs in that depth generation is embedded in a unified model that can first construct contexts including textual and visual ones before decoding the depth map.

\para{Textual Contexts (Depth Caption).}
Before estimating depth, we let the model explicitly reason about scene geometry and produce a concise textual summary of relative spatial structure (e.g., front/back ordering, occlusion relations, support/contact), which we term a ``depth caption''.
As a control, we also enable an unrestricted \texttt{think} mode to produce detailed, general-purpose image captions.
As shown in \Cref{tab:3d_rec} that geometry-focused textual context (the depth caption) improves depth estimation quality, whereas generic detailed captions provide little to no benefit. This suggests that context unrolling is most effective when the intermediate context is task-relevant and constraint-like, rather than merely verbose.

\para{Visual Contexts (Visual Tokens).}
Depth estimation remains a visual prediction problem in our framework. We therefore further augment the conditioning signal with self-rollout visual tokens, which tend to encode structured information (objectness, layout, coarse geometry cues) more explicitly. Injecting these tokens as visual context stabilizes the depth generation process and yields sharper, more globally consistent predictions.

We visualize depth estimation results with different contexts in \Cref{fig:depth_ablation}. For example, without any context, the model fails to properly separate the plant's leaves (the first column) and misses the ceiling lamps entirely (the third column). Depth captions help identify these objects, but ignore the depth difference between the left and right part of the wall, or incorrectly place the emitted light at the same depth plane as the lamp fixtures. Visual token context further corrects the geometry by distinguishing the depth difference between the wall's different parts and assigning illuminated areas depths similar to the background wall.

Overall, these results suggest, in a unified model, multimodal understanding and structured visual rollouts are not auxiliary add-ons. They function as context-producing primitives that constrain and guide 3D geometry, turning depth estimation from direct regression ($I \rightarrow D$) into context-conditioned inference ($D = \mathrm{Depth}(I \mid C_{\text{text}}, C_{\text{vis}})$).

\subsection{Discussion}

Our results consistently support a single conclusion across understanding, generation, 3D geometry, and 3D spatial reasoning: the primary value of a unified model is not capability aggregation, but unified context unrolling. Once multiple modalities and 3D operators are trained within one model, each task becomes an atomic primitive that can be invoked to construct intermediate, task-relevant context: ``text-think'' for constraint extraction, `` visual tokens'' for structural scaffolding, ``camera prediction'' and ``novel view synthesis'' for geometry-grounded verification. The final prediction is thus better viewed as context-conditioned inference rather than a direct mapping.

While our experiments focus on supervised/standard evaluations, unified context unrolling also suggests a plausible interaction with post-training (\textit{e.g.,} RL-style optimization), which we leave as future work. Intuitively, unifying more primitives enlarges the model decision space at inference time: the system could choose whether to answer directly or to first allocate compute to intermediate steps such as text-think, rolling out visual tokens, predicting camera pose, or running novel view synthesis as verification. In this view, post-training may be able to learn a policy over when and how to construct context, potentially improving robustness by adapting the context-building strategy to input difficulty and domain shift. More broadly, these considerations point to multimodal chain-of-thought / multimodal context construction as a promising research direction: instead of treating intermediate reasoning as purely textual, future systems may benefit from reasoning trajectories that interleave text, visual structure, and geometry-aware synthesis i.e., thinking with multiple modalities to build actionable context before decoding the final output.
\section{Evaluation}
In this section, we evaluate the performance of \method on multiple benchmarks, including multimodal understanding, image generation (\textit{i.e.,} text-to-image generation, image editing), video generation (\textit{i.e.,} text-to-video generation, video editing), 3D reconstruction (\textit{i.e.,} camera pose estimation and depth estimation). 
\begin{table}[t]
\centering
\caption{\textbf{Multimodal Understanding.} Performances on the standard benchmarks are reported, compared to other approaches with similar MoE architecture, activations and inference scheme (\textit{i.e.,} no-thinking).}
\label{tab:vlm_eval}
\resizebox{0.8\textwidth}{!}{
\renewcommand{\arraystretch}{1.0}
\resizebox{\textwidth}{!}{%
\begin{tabular}{l|c|c|c}
\toprule
 & Qwen3-VL-30B-A3B-Instruct~\cite{Qwen3-VL} & InternVL3.5-30B-A3B~\cite{wang2025internvl3} & \makebox[4cm]{\centering \method}\\
\midrule

BLINK~\cite{fu2024blink} & \textbf{67.7} & 60.4 & 63.0 \\
MMStar~\cite{chen2024we} & \textbf{78.4} & 72.0 & 63.8 \\
MMBench-v11~\cite{liu2024mmbench} & 78.4 & \textbf{84.8} & 75.3 \\
VlmsAreBlind~\cite{rahmanzadehgervi2024vision} & 67.5 & - & \textbf{76.4} \\
SimpleVQA~\cite{cheng2025simplevqa} & 52.7 & - & \textbf{53.3} \\
RealWorldQA~\cite{xai2024realworldqa} & 73.7 & 72.3 & \textbf{76.0} \\
Textvqa~\cite{singh2019towards} & - & 80.5 & \textbf{81.0} \\
AI2D~\cite{kembhavi2016diagram} & 85.0 & 86.8 & \textbf{91.5} \\
Chartqa~\cite{masry2022chartqa} & 86.8 & \textbf{87.4} & 86.9 \\
Docvqa~\cite{mathew2021docvqa} & \textbf{95.0} & 94.2 & 92.8 \\
HallusionBench~\cite{guan2024hallusionbench} & 61.5 & 53.8 & \textbf{70.1} \\
MuirBench~\cite{wang2024muirbench} & \textbf{73.0} & 53.1 & 64.2 \\
Erqa~\cite{team2025gemini} & \textbf{51.3} & 41.5 & 45.0 \\
MMSI-Bench~\cite{yang2025mmsi} & 30.3 & 27.5 & \textbf{31.5} \\
MVBench~\cite{li2024mvbench} & \textbf{72.3} & 72.1 & 68.4 \\
Video-MME$_{w/o sub.}$~\cite{fu2025video}& \textbf{74.5} & 68.7 & 67.2 \\

\bottomrule
\end{tabular}
}
}
\end{table}

\begin{table}[t]
\centering
\caption{\textbf{Image Generation Evaluation.} We validate the text-to-image and image editing on multiple open benchmarks, compared against various expertise approaches. }
\label{tab:t2i_eval}
\renewcommand{\arraystretch}{1.0}
\resizebox{\textwidth}{!}{%
\begin{tabular}{l|ccccc}
\toprule
\textbf{Models} & \textbf{GenEval2$\uparrow$} & \textbf{DPG$\uparrow$} & \textbf{LongText-EN$\uparrow$} & \textbf{LongText-CN$\uparrow$} &\textbf{Inhouse$\uparrow$} \\
\midrule
Qwen-Image~\cite{wu2025qwen} & 30.67 & 88.32 & 94.3 & 94.6 & 55.16\\
Z-Image~\cite{cai2025z} & 41.83 & 88.14 & 93.5 & 93.6 & 55.19\\
Flux~\cite{flux2} & 34.59 & 83.84 & 60.7 & 0.5 & 49.91\\
\midrule
\method & \textbf{54.12} & \textbf{88.55} & \textbf{97.5} & \textbf{96.8} & \textbf{63.87} \\

\bottomrule
\end{tabular}
\begin{tabular}{l|ccc}
\toprule
\multirow{2}{*}{\textbf{Models}}
& \multicolumn{3}{c}{\textbf{GEdit-Bench-EN (Full set)$\uparrow$}} \\
\cmidrule(lr){2-4}
& G\_SC & G\_PQ & G\_O \\
\midrule
Flux-Kontext-dev~\cite{flux_kontext}          & 7.16 & 7.37 & 6.51  \\
Step1X-Edit-v1.1~\cite{liu2025step1x}     & 7.66 & 7.35 & 6.97 \\
Step1X-Edit-v1.2~\cite{liu2025step1x}     & 7.77 & 7.65 & 7.24 \\
Emu-3.5~\cite{cui2025emu3} & 8.11 & 7.70 & \underline{7.59} \\ 
Z-Image-Edit~\cite{cai2025z} & 8.11 & 7.72 & 7.57 \\ 
Qwen-Image-Edit~\cite{wu2025qwen} & \underline{8.15} & \textbf{7.86} & 7.54  \\ 
\midrule
\method & \textbf{8.42} & \underline{7.85} & \textbf{7.75}  \\
\bottomrule
\end{tabular}
}
\end{table}
\subsection{Multimodal Understanding}

Given that \method is built upon an MoE architecture that usually contains more parameters yet less activations than prior unified models, we focus our comparison on VLMs at the similar scale. We therefore include Qwen3-VL-30B-A3B-Instruct~\cite{Qwen3-VL} and InternVL3.5-30B-A3B~\cite{wang2025internvl3} in our evaluation, as both are based on the same LLM backbone (i.e., Qwen3-30A3).
That said, strict one-to-one comparisons are inherently difficult, since differences in training data, optimization recipes, compute budgets, and vision encoders can all affect final performance. We summarize the visual understanding results in \Cref{tab:vlm_eval} to position our unified model among these representative baselines.
Without heavy post-training and distillation, \method achieves comparable performances across general VQA, chart and graph understanding, alignment, video and spatial understanding. 

\subsection{Image Generation}
We mainly evaluate the proposed method from two perspectives: image generation with text-only prompt or image-instruction pair. Current open-source image generators usually deliver two separate models for text-to-image and editing respectively. As a unified model, \method can naturally perform image generation and editing tasks, depending on the modality combination of input contexts. We thus report the performances on various benchmarks, comparing our proposed method with the expertise models.

\Cref{tab:t2i_eval} presents the main results on both text-to-image and image editing tasks, including GenEval2~\cite{kamath2025geneval}, DPG~\cite{hu2024ella}, LongText-EN~\cite{geng2025x}, Inhouse evaluation, and GEdit~\cite{liu2025step1x}.
Although prior approaches usually derive expertise models that focus on different tasks respectively, \method benefits from the MoE architecture and task unification, achieving comparable performances with only 3B activations.

\begin{table}[t]
\centering
\caption{\textbf{Video Generation Evaluation.} We validate the text-to-video on VBench1.0, compared against various expertise approaches. }
\label{tab:t2v_eval}
\resizebox{0.7\textwidth}{!}{
\renewcommand{\arraystretch}{0.5}
\begin{tabular}{l|ccc}
\toprule
\textbf{Models} & \textbf{Total Score$\uparrow$} & \textbf{Quality Score$\uparrow$} & \textbf{Semantic Score$\uparrow$} \\
\midrule
Wan2.1~\cite{wan2025wan} & \textbf{83.69} & \textbf{85.59} & 76.11 \\
Hunyuan Video~\cite{kong2412hunyuanvideo} & 83.43 & 85.07 & 76.88\\
\midrule
\method & 83.35 & 83.11 & \textbf{84.29} \\
\bottomrule
\end{tabular}
}
\end{table}
\begin{table}[t]
\centering
\caption{\textbf{Video Editing Results on FiVE Benchmark~\cite{li2025five}}.}
\label{tab:five}
\resizebox{\textwidth}{!}{%
\begin{tabular}{l c ccc cc c c ccccc}
\toprule
\multirow{2}{*}{\textbf{Method}}
& \multicolumn{1}{c}{\textbf{Structure}}
& \multicolumn{3}{c}{\textbf{Background Preservation}}
& \multicolumn{2}{c}{\textbf{Text Alignment}}
& \multicolumn{1}{c}{\textbf{Motion}}
& \multicolumn{5}{c}{\textbf{FiVE}} \\
\cmidrule(lr){2-2}\cmidrule(lr){3-5}\cmidrule(lr){6-7}\cmidrule(lr){8-8}\cmidrule(lr){9-13}
& Dist.$\times 10^{3}\downarrow$
& PSNR$\uparrow$
& LPIPS$\times 10^{3}\downarrow$
& SSIM$\times 10^{2}\uparrow$
& CLIPS.$\uparrow$
& CLIPS.$_{\mathrm{edit}}\uparrow$
& Fid S.$\times 10^{2}\uparrow$
& YN$\uparrow$
& MC$\uparrow$
& $\cup$$\uparrow$
& $\cap$$\uparrow$
& Acc$\uparrow$ \\
\midrule
Source Videos
& 0.00 & $\infty$ & 0.00 & 100.00 & 24.59 & 19.87 & 93.76 & -- & -- & -- & -- & -- \\
\midrule
TokenFlow~\cite{geyer2023tokenflow}
& 35.62 & 19.06 & 263.61 & 72.51 & 26.46 & 21.15 & 89.00
& 19.36 & 35.51 & 36.68 & 18.18 & 27.43 \\
DMT~\cite{yatim2024space}
& 85.95 & 14.71 & 404.60 & 51.64 & 26.66 & \textbf{21.44} & 82.30 
& 34.78 & \underline{62.06} & \underline{62.98} & 33.86 & \underline{48.42} \\
VidToMe~\cite{li2024vidtome}
& 22.37 & 21.15 & 263.91 & 70.69 & \underline{26.84} & 21.05 & \textbf{90.06} 
& 20.03 & 33.50 & 36.20 & 17.34 & 26.77 \\
AnyV2V~\cite{ku2024anyv2v}
& 71.36 & 15.90 & 348.59 & 50.77 & 24.89 & 19.72 & 60.36
& 30.62 & 45.42 & 48.96 & 27.09 & 38.02 \\
VideoGrain~\cite{yang2025videograin}
& \textbf{12.40} & \textbf{27.05} & \underline{185.21} & \underline{79.13} & 25.69 & 20.31 & 88.57
& 30.50 & 43.97 & 44.30 & 30.17 & 37.23 \\
Pyramid-Edit~\cite{li2025five}
& 28.65 & 20.84 & 276.59 & 71.72 & 26.82 & 20.20 & 80.59
& 33.67 & 54.01 & 56.36 & 31.31 & 43.84 \\
Wan-Edit~\cite{li2025five}
& \underline{12.53} & \underline{25.57} & \textbf{94.61} & \textbf{82.55} & 26.39 & \underline{21.23} & \underline{89.43}
& \underline{41.41} & 52.53 & 55.72 & \underline{38.22} & 46.97 \\
\midrule
\method
& 34.94 & 22.95 & 217.55 & 73.78 & \textbf{26.92} & 21.19 & 84.22
& \textbf{62.83} & \textbf{81.81} & \textbf{84.33} & \textbf{60.23} & \textbf{72.41} \\
\bottomrule
\end{tabular}%
}
\end{table}

\subsection{Video Generation}

Beyond image generation, \method can also synthesize videos with various combinations of multimodal instructions. Similarly, we report the performances on general text-to-video generation and video editing on widely used benchmarks (\textit{i.e.,} VBench~\cite{huang2023vbench} and FiVE~\cite{li2025five}.) 
\Cref{tab:t2v_eval} presents the text-to-video results on VBench where the proposed method achieves comparable performances. However, the current \method can only produce videos with the resolution of $480\times640$ and the duration of 12 seconds, which is far behind the state-of-the-art video generation expertise models. We believe these shortcomings would be weakened as further scaling up. Meanwhile, \Cref{tab:five} compares the video editing performances at the similar resolution and duration. Clearly, our method shows the significant superiority over other approaches in instruction following.


\begin{table*}[h!]
\centering
\caption{\textbf{Camera Pose Estimation on RealEstate10K and CO3Dv2.}}
\label{tab:cam_pose}
\begin{tabular}{l ccc ccc}
\toprule
\multirow{2}{*}{\textbf{Method}}
& \multicolumn{3}{c}{\textbf{RealEstate10K}}
& \multicolumn{3}{c}{\textbf{CO3Dv2}} \\
\cmidrule(lr){2-4}\cmidrule(lr){5-7}
& AUC$@$30 $\uparrow$
& RPE trans $\downarrow$
& RPE rot $\downarrow$
& AUC$@$30 $\uparrow$
& RPE trans $\downarrow$
& RPE rot $\downarrow$ \\
\midrule
Flare~\cite{zhang2025flarefeedforwardgeometryappearance}
& 84.42 & 0.4215 & 0.0532 & 72.23 & 2.1242 & 0.0342 \\
Cut3r~\cite{wang2025continuous}
& 85.32 & 0.4023 & 0.0424 & \underline{75.62} & \underline{1.5321} & 0.0331 \\
VGGT~\cite{wang2025vggt}
& \underline{88.23} & \underline{0.3886} & \underline{0.0386} & \textbf{86.23} & \textbf{1.1432} & \underline{0.0285} \\
\midrule
\method
& \textbf{88.32} & \textbf{0.3766} & \textbf{0.0289} & 75.21 & 1.5955 & \textbf{0.0269} \\
\bottomrule
\end{tabular}%
\end{table*}

\begin{table*}[h!]
\centering
\caption{\textbf{Monocular Depth Estimation.}}
\label{tab:monodepth_comp}
\resizebox{\textwidth}{!}{%
\begin{tabular}{l cc cc cc cc cc cc}
\toprule
\multirow{2}{*}{\textbf{Method}}
& \multicolumn{2}{c}{\textbf{NYU}}
& \multicolumn{2}{c}{\textbf{KITTI}}
& \multicolumn{2}{c}{\textbf{SINTEL}}
& \multicolumn{2}{c}{\textbf{ETH3D}}
& \multicolumn{2}{c}{\textbf{DIODE}} \\
\cmidrule(lr){2-3}\cmidrule(lr){4-5}\cmidrule(lr){6-7}\cmidrule(lr){8-9}\cmidrule(lr){10-11}
& $\delta_1\uparrow$
& AbsRel$\downarrow$
& $\delta_1\uparrow$
& AbsRel$\downarrow$
& $\delta_1\uparrow$
& AbsRel$\downarrow$
& $\delta_1\uparrow$
& AbsRel$\downarrow$
& $\delta_1\uparrow$
& AbsRel$\downarrow$ \\
\midrule
Marigold~\cite{ke2023repurposing}
& 92.75 & 0.0781 & 87.87 & 0.1108 & 62.24 & 0.4666 & 97.12 & 0.0564 & 81.64 & 0.2266 \\
Cut3r~\cite{wang2025continuous}
& 91.64 & 0.0824 & 86.42 & 0.1253 & 55.64 & 0.4723 & 95.34
& 0.0632 & 75.21 & 0.3521 \\
DA3 giant~\cite{depthanything3}
& 94.78 & 0.0579 & 93.96 & 0.0824 & \underline{66.54} & \underline{0.3821} & \underline{98.79}
& \underline{0.0324} & \underline{82.69} & \underline{0.2050} \\
VGGT~\cite{wang2025vggt}
& \underline{96.10} & \textbf{0.0499} & \underline{94.29} & \underline{0.0803} & 66.11 & 0.4551 & 98.35
& 0.0326 & 82.15 & 0.2115 \\
\midrule
\method
& \textbf{96.22} & \underline{0.0542} & \textbf{96.92} & \textbf{0.0621} & \textbf{74.27} & \textbf{0.3340} & \textbf{98.91}
& \textbf{0.0312} & \textbf{83.83} & \textbf{0.2034} \\
\bottomrule
\end{tabular}%
}
\end{table*}

\subsection{3D Geometry}
As recent feedforward models (\textit{i.e.,} VGGT) have already demonstrated the effectiveness of unifying multiple 3D tasks, \method also supports typical tasks in 3D vision: camera pose estimation and monocular depth estimation, which together define 3D correspondences. We therefore compare against the expertise model in the 3D field to anchor the capability of unified models.

\para{Camera Pose Estimation.}
As shown in \Cref{tab:cam_pose}, we evaluate our method on camera pose estimation against Flare, Cut3r, and VGGT on the RealEstate10K and CO3Dv2 datasets. On RealEstate10K, our method achieves state-of-the-art performance, surpassing all baselines across all three metrics. On the object-centric CO3Dv2 benchmark, our approach secures the best result in translation error, although other metrics are not as competitive. We attribute this discrepancy to our data collection process, which may not have sufficiently covered object-centric scenes. Although it is challenging to make a perfectly fair comparison, these results are significant: they demonstrate that a unified model, even without explicit 3D inductive biases and using only text to represent camera parameters, can achieve strong performance with appropriate training data. Such 3D capability can be regarded as one built-in context, revealing the potential in spatial understanding.

\para{Monocular Depth Estimation.}
We further validate our model on the monocular depth estimation task, with results benchmarked against specialist models on five standard datasets presented in \Cref{tab:monodepth_comp}. Our method demonstrates remarkable zero-shot performances on multiple benchmarks. These comprehensive results underscore our model's strong generalization capabilities, proving it can achieve or even exceed the performance of specialized, single-task models across a wide range of domains without task-specific fine-tuning.

\section{Acknowledgment}
We thank Shu Liu, Xuejiao Zeng, Xiaojie Li, Renfei Sun, Ashley Kim, Ruoqing Hu, Xi Lin, Liyang Liu, Xinyu Zhang, Liang Li, Shuangye Li, Yuhong Yang, Hongxiang Hao, Heng Zhang, Zanbo Wang, Lishu Luo, Sijin Wu, Faming Wu, Xudong Sun for their helpful contribution and discussion.

\clearpage

\bibliographystyle{plainnat}
\bibliography{main}




\end{document}